# Does constituency analysis enhance domain-specific pre-trained BERT models for relation extraction?

Anfu Tang[1,2], Louise Deléger[1], Robert Bossy[1], Pierre Zweigenbaum[2], Claire Nédellec[1]
[1]Université Paris-Saclay, INRAE, MaIAGE, 78350, Jouy-en-Josas, France
[2]Université Paris-Saclay, CNRS, Laboratoire interdisciplinaire des sciences du numérique, 91405, Orsay, France

*Abstract*—Recently many studies have been conducted on the topic of relation extraction. The DrugProt track at BioCreative VII provides a manually-annotated corpus for the purpose of the development and evaluation of relation extraction systems, in which interactions between chemicals and genes are studied. We describe the ensemble system that we used for our submission, which combines predictions of fine-tuned bioBERT, sciBERT and const-bioBERT models by majority voting. We specifically tested the contribution of syntactic information to relation extraction with BERT. We observed that adding constituent-based syntactic information to BERT improved precision, but decreased recall, since relations rarely seen in the train set were less likely to be predicted by BERT models in which the syntactic information is infused. Our code is available online[1].

*Keywords—Relation Extraction; Deep Learning; Transformers; BERT (Bidirectional Encoder Representations from Transformers); constituency analysis; chemical-gene interactions*

## I. Introduction

Identifying interactions between different chemicals and drugs is important for many biomedical applications, for example, creation of knowledge bases from the published literature (1). Though manually extracting these relations by experts is possible, establishing an automatic text mining system will help accelerate the processing and save time and resources. The BioCreative VII track 1 DrugProt task aims to evaluate the quality of relation-extraction systems that are able to identify chemical-drug interactions in the scientific literature. In this paper, we describe the methods that we used for our submission to the task: ensembles of fine-tuned bioBERT models and fine-tuned bioBERT models in which constituency information is infused (const-bioBERT). We release our code that we used during the challenge on Github[1].

## II. Data

### A. Corpus

Lack of high-quality annotated corpus has always been a point which makes biomedical text mining more challenging, compared to the general domain in which more annotated data is available. The BioCreative Challenge has been devoted to provide reliable sources for biomedical text mining. Similar to the ChemProt track (2) of BioCreative VI, in the DrugProt track (3), a corpus with chemical-gene relations annotated by the organisers is provided, which contains 3,500, 750, and 10,750 PubMed records, respectively in the training, development and testing sets. 17,070 and 3,731 relations are annotated respectively in the train and development set.

One specificity of the corpus is that an entity pair may be related by multiple labeled relations. 201 and 30 pairs linked by two relations are found in the train and development sets. Though the corpus contains cross-sentence relations, their proportion is small. Only 3 relations in the train set are found to span two consecutive sentences. Therefore, in this challenge we treated the problem as a multi-class classification task on intra-sentence relations.

### B. Data preparation

We first segmented sentences using Stanza (4), and mapped entities to sentences to which they belong depending on their spans; we then extracted and tagged every possible pair of entities (selected by entity types: CHEMICAL, GENE). In the two previous steps, errors are reported if: 1) a word is found to be split across two sentences; 2) a relation is found to be cross-sentence (on the train and development set). Reported errors were manually checked and 22 errors were found to be sentence segmentation errors and were then corrected. In the established dataset, for each pair of entities, we used the original sentence containing the two target arguments and added markers "@@" at the start and the end of the subject, while "$$" was used as the marker for the object. In cases where the subject and the object arguments overlap, the full span of the two arguments was selected and then enclosed within "¢¢" markers.

## III. Methods

### A. BERT: bidirectional encoder representations from transformers

Pre-trained language models based on the Transformer architecture (5) have been proved to be state-of-the-art on many NLP tasks over recent years. BERT is a contextualized word representation model based on bidirectional transformers (6). Pre-training of BERT includes two tasks: 1) Masked Language Modeling: predict randomly masked tokens in a sequence; 2) Next Sentence Prediction: predict whether two sentences are consecutive in the original document. The pre-training corpus includes the English Wikipedia and the BookCorpus, which in total contain 3.3 billion words.

---

1. https://github.com/Maple177/drugprot-relation-extraction

## B. Domain-specific BERT: bioBERT and sciBERT

Biomedical texts are usually more complicated than texts in the general domain in the aspect of vocabulary and sentence structure, thus several pre-trained variants of BERT are proposed with the expectation that using a domain-specific corpus or vocabulary can help improve the performance of BERT. In this challenge, we chose bioBERT (7) and sciBERT (8), two pre-trained BERT variants that were shown to perform well on several public biomedical datasets. All pre-trained weights are loaded using the Python library provided by HuggingFace (9). The versions that we use are: bert-base-uncased, biobert-base-cased-v1.1, scibert_scivocab_uncased, respectively for BERT, bioBERT and sciBERT. Fine-tuning was performed using tagged sentences prepared as mentioned in the previous section.

## C. const-BERT: BERT integrating constituent information

In biomedical texts, many entities are compound nouns themselves or belong to a noun phrase. We hypothesize that constituency analysis may help simplify the syntax of sentences and thus render the relation between entities more explicit. Therefore, to achieve this goal, we propose a novel BERT-based neural network that integrates constituent information, to distinguish the novel model from the vanilla BERT, we name these proposed models by adding a prefix "const-" before the name of BERT variant that is used, e.g. const-BERT, const-bioBERT, etc. (Fig. 1).

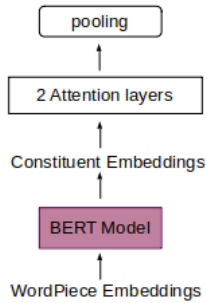

Fig. 1. Overview of the const-BERT model

We first apply a syntactic parser to each input sentence to produce a constituency parse tree. Then in the proposed neural network, tokens are firstly segmented into WordPieces (10) and mapped to WordPiece embeddings using pre-trained WordPiece tokenizers provided by corresponding BERT variants, then passed through a pre-trained BERT model. We group together WordPiece embeddings in the output of the BERT model into constituent embeddings according to the constituency parse tree using the following rule: 1) for each token, sum all its word pieces to obtain the token embedding; 2) traverse the constituency tree depth first to identify NPs and VPs that contain no NP or VP; compute an embedding for each such constituent by summing its token embeddings. Our method can be viewed as simple chunking where only NPs or VPs are included in chunk rules. Constituency parse trees are obtained from the Berkeley Neural Parser (11). After the pre-trained BERT model, we use two extra Attention (5) layers to contextualize constituent embeddings; the output is then fed into a pooling layer as in most standard fine-tuning neural network pipelines.

## D. A Majority Voting System

We fine-tuned three pre-trained models: bioBERT, sciBERT and const-bioBERT on the train set and used the F1-score on the development set as the criterion for early stopping. Hyperparameters are kept the same for all experiments: We apply standard dropout of 0.1 on Attention weights of all layers, and standard dropout of 0.1 on all intermediate layer outputs. We use mini-batch training with batch size of 16 and constant learning rate of 2e-5. All parameters were trained using the Adam Algorithm (12) to optimize the mean of the weighted binary cross entropy over all relation types, i.e. assuming there are K relation types, for a certain relation type $r$, given the raw outputs of neural network $x = \{x_1,…,x_K\}$, and the labels $y = \{y_1, … ,y_K\}$, the loss for the relation $r$ is:

$$l_r = -w_r[\, y_r \log(x_r) + (1 - y_r) \log(1 - x_r)] \quad (1)$$

$$w_r = \frac{\sum_{i=1}^{K} N_i}{N_r} \quad (2)$$

where $N_r$ denotes the number of examples labelled with relation $r$.

For each pre-trained model, 8 models were trained using different randomly initialized weights, thus in the end 24 models were trained. We built 5 ensembles with 5 different model combinations for two reasons: 1) voting in an ensemble containing a single type of model can help stabilise the output and mitigate the influence of random errors due to different random initialisation of model weights and majority voting; 2) we suppose that an ensemble containing models of different BERT variants may counteract the system errors of each other. We used the majority vote to combine the model results within an ensemble. In cases where two labels share the maximum count value for a certain example, the example was predicted with these two labels.

## IV. EXPERIMENTAL RESULTS & ANALYSIS

We built 5 ensembles using respectively:
1) 7 bioBERT models
2) 4 bioBERT models and 4 sciBERT models
3) 7 const-bioBERT models
4) 4 bioBERT models and 4 const-bioBERT models
5) 4 bioBERT models, 4 sciBERT models and 4 const-bioBERT models

For 1) and 3), from the 8 initially trained models, we removed the one with the lowest F1-score on the development set; for 2), 4) and 5), the 4 models that we selected for each BERT variant were the 4 best models out of 8 according to their F1-scores on the development set. The architecture of our system is shown below (Fig. 2).

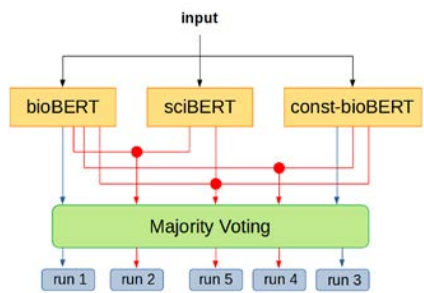

Fig. 2. Architecture of the system for DrugProt

The DrugProt test set contains 10,750 abstracts, among which a subset of 750 abstracts is used for evaluation. Table 1 shows the micro-precision, recall and F1-score obtained by our system on the test subset as reported by the organisers.

The fact that Run 2 outperforms Run 1 verifies our hypothesis that different types of BERT variants in an ensemble may counteract system errors of each other. From the similar results of run 1 and run 4, we conclude that const-bioBERT is fine-tuned in the same way as bioBERT, since merging results of 8 bioBERT models and merging a combination of 4 bioBERT models and 4 const-bioBERT models show no significant difference.

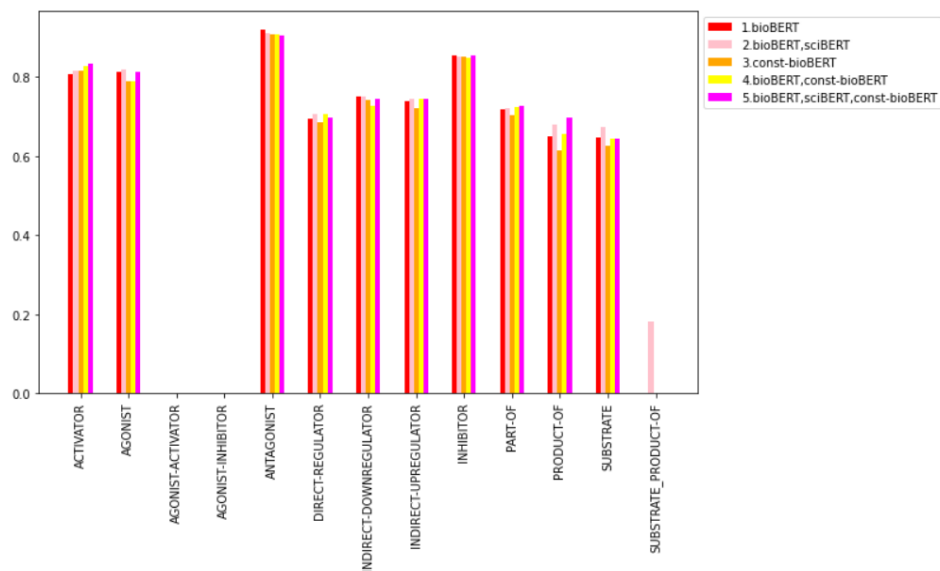

Fig. 3. results per relation on the test set of DrugProt

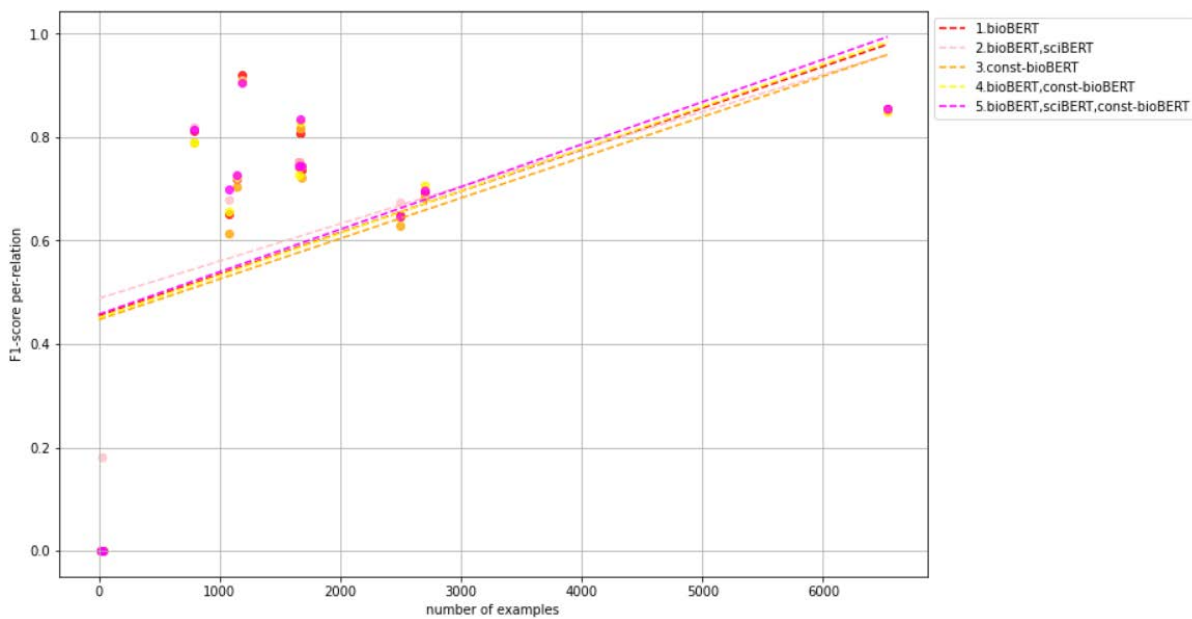

Fig. 4. number of examples in the train set of DrugProt per relation with respect to corresponding F1-scores on the test set

TABLE I. RESULTS ON THE TEST SET

| Run | System | Precision | Recall | F1-score |
|---|---|---|---|---|
| 1 | bioBERT | 0.7571 | 0.7797 | 0.7682 |
| 2 | bioBERT sciBERT | 0.7546 | **0.7966** | **0.7750** |
| 3 | const-bioBERT | 0.7604 | 0.7622 | 0.7613 |
| 4 | bioBERT const-bioBERT | 0.7542 | 0.7834 | 0.7685 |
| 5 | bioBERT sciBERT const-bioBERT | **0.7657** | 0.7845 | **0.7750** |

The best result is in bold. The next best result(s) is underlined.

Detailed results for each relation are reported by the organizers as well and we summarise them in Fig. 3. We first seek to investigate the influence of the number of training examples. A natural hypothesis is that the more training examples of a certain relation exist, the better our system is able to predict for this type of relation. To test this hypothesis, for each BERT variant, we plot 13 points in the 2-D space for each type of relation, where the horizontal coordinate of each point denotes the number of training examples for the corresponding type of relation and the vertical coordinate denotes the corresponding F1-score. Then for each BERT variant, a dashed line is plotted which minimises the squared error as in Fig. 4.

There are three main observations: 1) when sufficient (>500) training examples exist, the performance for a certain type of relation is no longer limited by the number of training examples; 2) the impact of the number of training examples for different relations is similar among different BERT variants; 3) our system scores 0 F1-score for relations with too few examples: AGONIST-INHIBITOR, SUBSTRATE_PRODUCT-OF, AGONIST-ACTIVATOR with 15, 27, 39 examples in the train set respectively (though the second ensemble scores 0.18 for the relation SUBSTRATE_PRODUCT-OF, it is significantly much lower than scores for relations with more than 500 examples, we include it in the discussion as well). To figure out the cause of null performances on the three relations, we list below the number of examples predicted as one of the three relations in our predictions:

TABLE II. NUMBER OF PREDICTIONS FOR THE THREE RELATIONS WITH 0 F1-SCORE

| Run | AGONIST-INHIBITOR | SUBSTRATE_PRODUCT-OF | AGONIST-ACTIVATOR |
|---|---|---|---|
| 1 | 0 | 0 | 128 |
| 2 | 3 | 4 | 138 |
| 3 | 0 | 0 | 0 |
| 4 | 0 | 0 | 6 |
| 5 | 0 | 2 | 26 |

We obtain 0 F1-score on the three above relations for different reasons for all 5 runs. For ensembles in which the majority is BERT variants without syntax, some examples are predicted as one of the three relations but in fact all these predictions are wrong: in a word for these ensembles the null performance is caused by wrong positive predictions. In contrast, for ensembles that contain BERT variants with syntax, the null performance is due to no positive predictions. It is also interesting to find that adding constituency information seems to make the prediction less likely, i.e. there is no prediction for these relations that are rarely seen during training. We conjecture that it is due to the extra Attention layers which enhance the fine-tuning but reduce the impact of pre-trained weights.

Agonist-activator and Agonist inhibitor are two subtypes of Agonist with very few occurrences. A possible direction to handle them would be as a sub-case of Agonist, with a second-level classifier. A similar direction might be followed for Substrate-product-of, which is a sub-type of both Substrate and Product-of.

V. CONCLUSION

In this manuscript we described our submission to the BioCreative VII DRUGPROT task and analysed the reported results on the test set. Though adding constituency information brings no improvement according to the experimental results, it is found that ensembles that integrate syntactic information perform differently compared to original BERT variants. Further studies will be conducted by investigating concrete examples on which original BERT variants make mistakes that are corrected by adding syntax.